\pdfoutput=1

\documentclass[11pt]{article}

\usepackage{EMNLP2022}

\usepackage{times}
\usepackage{latexsym}
\usepackage{soul}

\usepackage[T1]{fontenc}
\usepackage[utf8]{inputenc}

\usepackage{microtype}

\usepackage{inconsolata}

\usepackage{multirow}
\usepackage{multicol}
\usepackage{amssymb}%
\usepackage{pifont}%
\usepackage{booktabs}
\usepackage{graphics}
\usepackage{graphicx}
\usepackage{tablefootnote}
\usepackage{footnote}
\usepackage{soul}
\usepackage{fdsymbol}

 \newcommand{\sm}[1]{\textcolor{black}{#1}}
\newcommand{\changejvj}[1]{\textcolor{black}{#1}}
\newcommand{\changemrungta}[1]{\textcolor{black}{#1}}

\title{Geographic Citation Gaps in NLP Research}

\makeatletter
\def\@fnsymbol#1{\ensuremath{\ifcase#1\or \clubsuit\or \clubsuit\or
\mathsection\or \mathparagraph\or \|\or **\or \clubsuit\clubsuit
\or \clubsuit\clubsuit \else\@ctrerr\fi}}
\makeatother

\author{
 {\hypersetup{linkcolor=black} Mukund Rungta$^{\diamondsuit}$\thanks{\ \ \ Equal contribution.} \;, Janvijay Singh$^{\diamondsuit\clubsuit}$, Saif M. Mohammad$^{\heartsuit}$, Diyi Yang$^{\lozenge}$}\\
$^{\diamondsuit}$ {School of Interactive Computing, Georgia Institute of Technology} \\
$^{\heartsuit}$ {National Research Council Canada} \\
$^{\lozenge}$ {Stanford University} \\
 \textcolor{darkblue}{\texttt{\{\href{mailto:mrungta8@gatech.edu}{mrungta8}, \href{mailto:iamjanvijay@gatech.edu}{iamjanvijay}\}@gatech.edu}} \\
 \texttt{\href{mailto:saif.mohammad@nrc-cnrc.gc.ca}{saif.mohammad@nrc-cnrc.gc.ca}} \\
  \texttt{\href{mailto:diyiy@stanford.edu}{diyiy@cs.stanford.edu}} \\
}

\begin{document}
\maketitle

\begin{abstract}
In a fair world, people have equitable opportunities to education, to conduct scientific research, to publish, and to get credit for their work, regardless of where they live. However, it is common knowledge \sm{among researchers} that a vast number of papers accepted at top NLP venues come from 
a handful of western countries
and (lately) China; whereas, very few papers from Africa and South America get published. %
Similar disparities are also believed  to exist for paper citation counts.
In the spirit of ``\emph{what we do not measure, we cannot improve}'', this work asks a series of questions on the relationship between geographical location and publication success (acceptance in top NLP venues and citation impact). We first created a dataset of 70,000 papers from the ACL Anthology, extracted their meta-information, and generated their citation network. We then %
show that not only are there substantial geographical disparities 
in paper acceptance and citation
but also that these disparities persist even when controlling for a number of variables such as venue of publication and sub-field of NLP. Further, despite some steps taken by the NLP community to improve geographical diversity, we show that the disparity in publication metrics across locations is still on an increasing trend since the early 2000s.
We release our code and dataset here: \url{https://github.com/iamjanvijay/acl-cite-net}.

\end{abstract}

\section{Introduction}
\sm{Progress in science is accelerated by a sharing of ideas. 
However, there have been numerous instances in history where the predominance of one group of people in science, and the silencing
of others, has led to the publication of harmful pseudo-science \cite{gould1996mismeasure,saini2019superior}.
Particularly egregious examples include the publication of theories and ideas on racial hierarchy \cite{plutzer2013racial}, male superiority \cite{huang2020historical}, gender binary \cite{darwin2017doing}, and eugenics  \cite{10.1093/jsh/shv053}. 
It has also been shown that a lack of inclusion in invention and discovery leads to fewer technologies for the excluded group.
For example, \citet{koning2021we} show how fewer technologies and health products are designed for women and
\citet{Bender_2011}, \citet{bird-2020-decolonising} and \citet{mohammad2019nlpscholar} show how a number of language technologies are designed for only a small number of languages.}

\sm{In this paper, we explore geographic inclusion in Natural Language Processing (NLP) research. Our premise is that}
in a fair world, people have equitable opportunities to education, to conduct scientific research, and to publish, regardless of where they live. 
\sm{However, researchers in the field know that} a vast number of papers accepted at top NLP conferences and journals come from 
a handful of western countries
and (lately) China. On the other hand, very few papers with African and South American authors are published. 

\sm{Further, the papers that get a majority of citations tend to be from a small number of institutions. %
Highly funded universities and research labs also tend to garner greater early visibility for their papers. 
Some of these papers might be cited more simply because the affiliate university or lab is perceived as prestigious \citep{amara2015can, hurley2013deconstructing}.} 
\citet{price1965networks} examined the growth of citation networks and showed that 
papers with more early citations are likely to be cited more in the future
(the ``\emph{rich get richer}'' phenomenon). %

Citations received by a research article serve as one of the key quantitative metrics to estimate its impact. %
Citations-based metrics, such as h-index \citep{hirsch2005index, bornmann2009state}, can have a considerable impact on a researcher’s career, funding received, and future research collaborations. Citation metrics are also commonly taken into consideration in determining university rankings and overall scientific outcomes from a country. 
Thus, the degree of equity in citations across geographic regions can act as one of the barometers of fairness in research.
Furthermore, geographic location directly correlates to the languages spoken in an area. Therefore, to increase the reach of NLP beyond high-resource languages, it is important to elevate the research pursued in languages from these under-represented regions.

In this work, we investigate the impact of a researcher's geographic location on their citability for the field of NLP. 
We examine tens of thousands of articles in the ACL Anthology (AA) (a digital repository of public domain NLP articles), and generate citation networks for these papers using information from 
Semantic Scholar,
to quantify and better understand disparities in citation based on the geographic location of a researcher. 
We consider a set of candidate factors that might impact citations received and perform both qualitative and quantitative analyses to better understand the degree to which they correlate with high citations. 

However, it should be noted that we do not explore the cause of citation disparities. Reasons behind such location-based disparities are often complex, inter-sectional, and difficult to disentangle. Through this work we aim at  bringing the attention of the community 
to geographic disparities in research. \sm{We hope that work in this direction} will inspire actionable steps to improve geographic inclusiveness and fairness in research.

\section{Dataset}
\label{"subsec:country_extract"}
As of January 2022, the ACL Anthology (AA) had 71,568 papers.\footnote{{\url{https://aclanthology.org/}}}
We extracted paper title, names of authors, year of publication, and
venue of publication for each of these papers from the repository. Further, we used information about the AA papers in Semantics Scholar\footnote{ {\url{https://www.semanticscholar.org/}}} to identify which AA papers cite which other AA papers --- \textit{the AA citation network}.  
Since the meta-information of the papers in AA and Semantic Scholar does not include the affiliation or location of the authors, we developed a simple heuristic-based approach to obtain %
\sm{affiliation}
information from the text of the paper. 

We refer to our dataset as the {\it AA Citation Corpus}. It includes the AA citation graph, author names, unique author ids (retrieved from Semantic Scholar), conference or workshop title, month and year of publication, and country associated with the author's affiliation. \textit{We make the AA Citation Corpus freely available.}

Detailed steps in the construction of the citation network and the extraction of \sm{affiliated} country information are described in the subsections below. 

\subsection{Citation Graph Construction} 
To create the citation graph, we collected the BibTeX entries of all 
the papers in the anthology. \changemrungta{We filtered out the entries which were not truly research papers such as  forewords, prefaces, programs, schedules, indexes, invited talks, appendices, session information, newsletters, lists of proceedings, etc. } Next, we used Semantic Scholar APIs\footnote{\url{https://www.semanticscholar.org/product/api}} to identify unique Semantic Scholar ID (SSID) corresponding to each paper in the BibTeX. For this, we queried the Semantic Scholar APIs in two ways: (a) Using the ACL ID present in BibTeX, which ensures that correct SSID was retrieved for a paper in BibTeX; and (b) for papers whose SSID cannot be retrieved using ACL ID, we searched the paper using the paper title mentioned in BibTeX. In (b), to ensure correctness of the retrieved SSID, we take the fuzzy string matching score\footnote{\url{https://pypi.org/project/fuzzywuzzy}} between title in BibTeX and that retrieved from Semantic Scholar. SSIDs with fuzzy score greater than 85\% are marked as correct. For the remaining retrieved SSIDs, we manually compared the title in BibTeX and the one retrieved from Semantic Scholar. 

We were able to retrieve correct SSIDs for \textbf{98.63\%} of the papers in the ACL Anthology. %
Finally, we queried the Semantic Scholar APIs with the SSIDs 
of each of the AA papers
to retrieve the SSIDs of the 
papers cited in the AA papers.
With this information, we created the AA citation graph. 

\subsection{Country Information Extraction}
We inferred the authors' affiliated country from the textual information extracted from the research paper PDFs.
We used SciPDF Parser\footnote{\url{https://github.com/titipata/scipdf_parser}}, a python parser for scientific PDF, to extract text from any PDF. Section-based parsing by this tool helps us to concentrate only on the header, which contains information about the author's affiliation. \sm{The considerable} differences in templates of papers published across different venues and years 
presented several challenges.
\sm{We first compiled an exhaustive list of countries and their universities from the web. (Details in Appendix \ref{"app:loc_extraction"}.)}
For each paper, 
\sm {we examine the affiliation section to identify mentions of a country (using our list of countries).}\footnote{Even if an author has multiple affiliations (countries) we only consider the ones mentioned in the paper.} 
Using this approach we were able to map each paper with its affiliated country. Table \ref{tab:country_stat} shows the number of papers having n-country tags, where $n = \{0, 1, >$1$\}$ represents no country, one country and multiple countries respectively. Further, as the mapping of paper to the country was automatically constructed, the authors manually annotated the ground truth country tag for 1000 papers. This was done to analyze the correctness of the automatically identified country tags. These papers were selected at random from the dataset. Out of 1000 papers, country tags for 845 (\textbf{84.5\%}) exactly match the ground truth. For most of the remaining unmatched cases, the algorithm either missed one country from the list or was unable to find any country tag for the paper.

\begin{table}[t!]
\begin{center}
\setlength{\tabcolsep}{5pt}%
\begin{tabular}{rr}
\toprule
\# Countries & Number of papers \\

\midrule

0: no country & 14,818 \\ 
1: one country & 48,815 \\ 
$>$1: multiple countries & 7,062 \\

\bottomrule 
\end{tabular}
\end{center}
\caption{\label{tab:country_stat} Count of papers \sm{by the number of automatically} inferred affiliated countries.
}

\end{table}

\begin{figure*}[!ht]
    \centering
    \includegraphics[width=2\columnwidth,trim=0 20 0 17.5,clip]{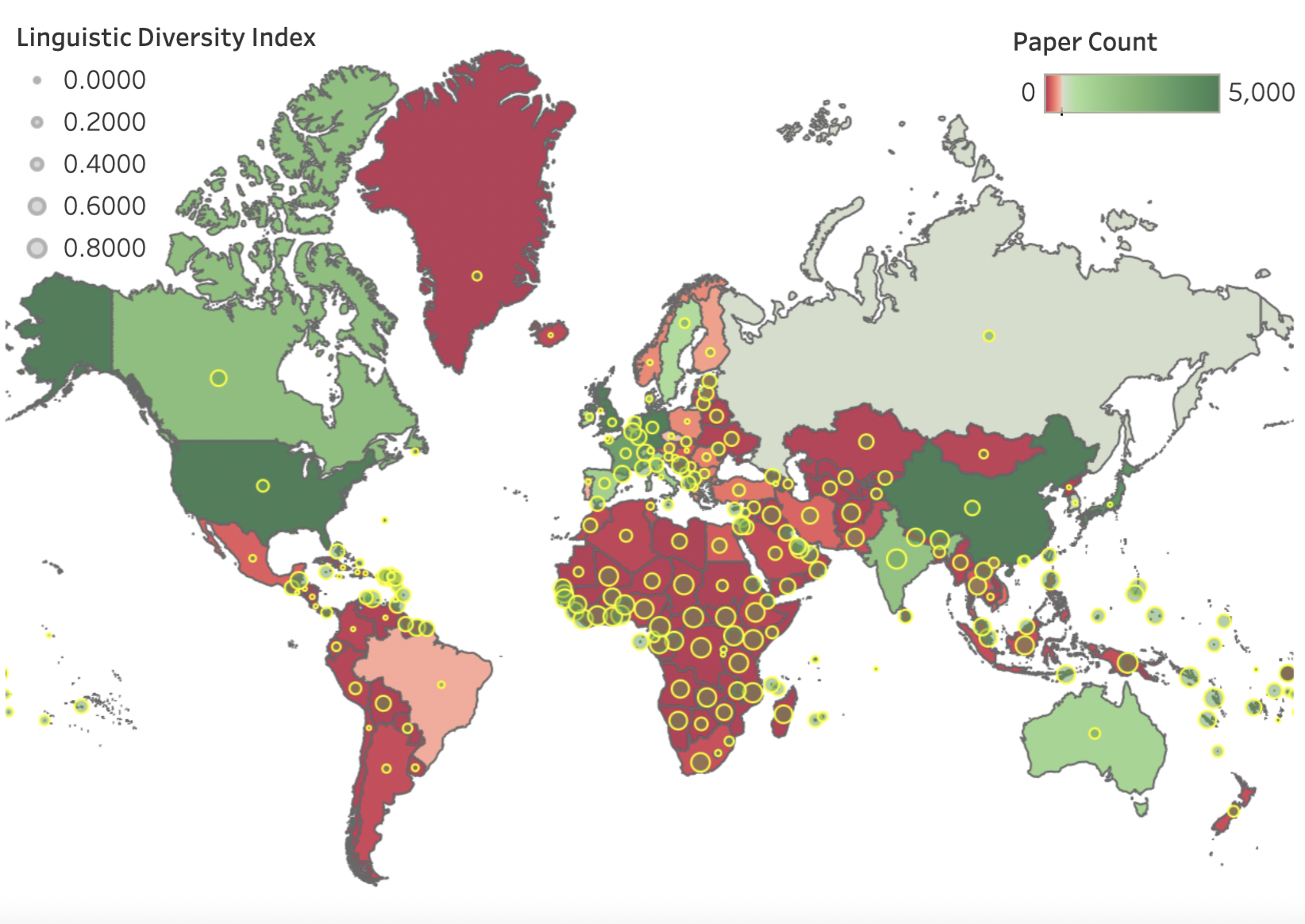} %
    \caption{World-map showing variation of linguistic diversity (shown through circles of different sizes) and volume of NLP research (shown through color shades from red to green) across the globe.} 
    \label{fig:world_map}
\end{figure*}

\section{Disparity in Citation based on Location}

We use the \textit{AA citation Corpus} to answer a series of questions on disparity of \sm{publications and} citations across geographic location. 
We start with a look at the number of publications from around the world, followed by an examination of their citations.

\vspace{2mm}
\noindent
\textbf{Q1. Is there a disparity in the number of NLP publications across different countries?
How does %
\sm{the amount of publications}
correlate with 
linguistic diversity?}\\ %

\noindent %
\textbf{A.} \sm{We used counts of papers from the AA Citation Corpus to determine
the number of papers from each country, as visualized in Figure  \ref{fig:world_map}.}
\sm{For an even coarser view, we also examined a partition of the world into ten regions.}\footnote{One can partition the world map into regions in many ways. We made use of the partition provided by the United Nations Geo-scheme: \url{https://en.wikipedia.org/wiki/List_of_countries_by_United_Nations_geoscheme}. This list includes seventeen subregions, and we combine some of these subregions into \sm{ten} coarser regions for simplicity.} 
We calculated the total number of papers from each region by aggregating papers from all countries present in this region. We also aggregate citation counts 
of papers by region.

\paragraph{Discussion}
Figure  \ref{fig:world_map} shows 
huge disparities for the number of publications among countries. The western world which includes United States, Canada, United Kingdom, France, Germany, etc.\@ dominates the network with high publication count. On the other hand, most countries in Africa, South America, Eastern Europe, South East Asia, and Middle East remain in the red zone with very few publications till date. %
When %
examining language diversity\footnote{ {\url{https://en.wikipedia.org/wiki/Linguistic_diversity_index}}}
(indicated by size of yellow dot), we see that countries in the red zone have the highest language diversity. Higher linguistic diversity indicates larger number of different languages spoken in that geographic region.
\sm{For example, the list of countries with the highest
 number of languages includes: Indonesia (710), Nigeria (524), India (453), and Brazil (228).}\footnote{ {\url{https://en.wikipedia.org/wiki/Number_of_languages_by_country}}}
 
More work on these languages is needed, 
by local researchers in partnership with the language communities.
One recent effort in this regard is project Masakhane, a grassroots organisation whose mission is to strengthen and spur NLP research in African languages, for Africans, by Africans.\footnote{ {\url{https://www.masakhane.io/}}} 
This analysis showcases the huge disparity in the number of publication from each country. %
Through the questions ahead, we further uncover geographic patterns in citations, across these mid-tier and top-tier publishing countries.

\vspace{2mm}
\noindent
\textbf{Q2. How has the citation count ("\emph{influence}" of NLP research) of papers from different regions changed over the years?}\\

\noindent 
\textbf{A.} To study this question, we examine the following metric: mean citation count per paper for each country until certain year. Formally, this metric can be defined as follows:

$$M C_{(j, k)}=\frac{\sum_{i \in P_{k}} C_{k}(i) \mathbb{I}_{i \in j}}{\sum_{i \in P_{k}} \mathbb{I}_{i \in j}}$$

\noindent where $M C_{(j, k)}$ indicates mean-citation count of country-$j$ until year-$k$. $C_{k}(i)$ indicates citation count of paper-$i$ until year-$k$. $\mathbb{I}_{i \in j}$ is $1$ if paper-$i$ belongs to country-$j$ otherwise $0$. $P_k$ indicates the set of all the papers published until year-$k$. To calculate this metric, we create multiple citation networks each containing papers published only until a certain year. For instance, a citation graph associated with year-$2000$ will only contain papers published until $2000$ --- with papers as nodes and references as edges. Using such a citation network associated with year-$k$, we can then easily evaluate $M C_{(j, k)}$ for any country-$j$.

\begin{figure}[t!]
    \centering
    \includegraphics[width=1\columnwidth]{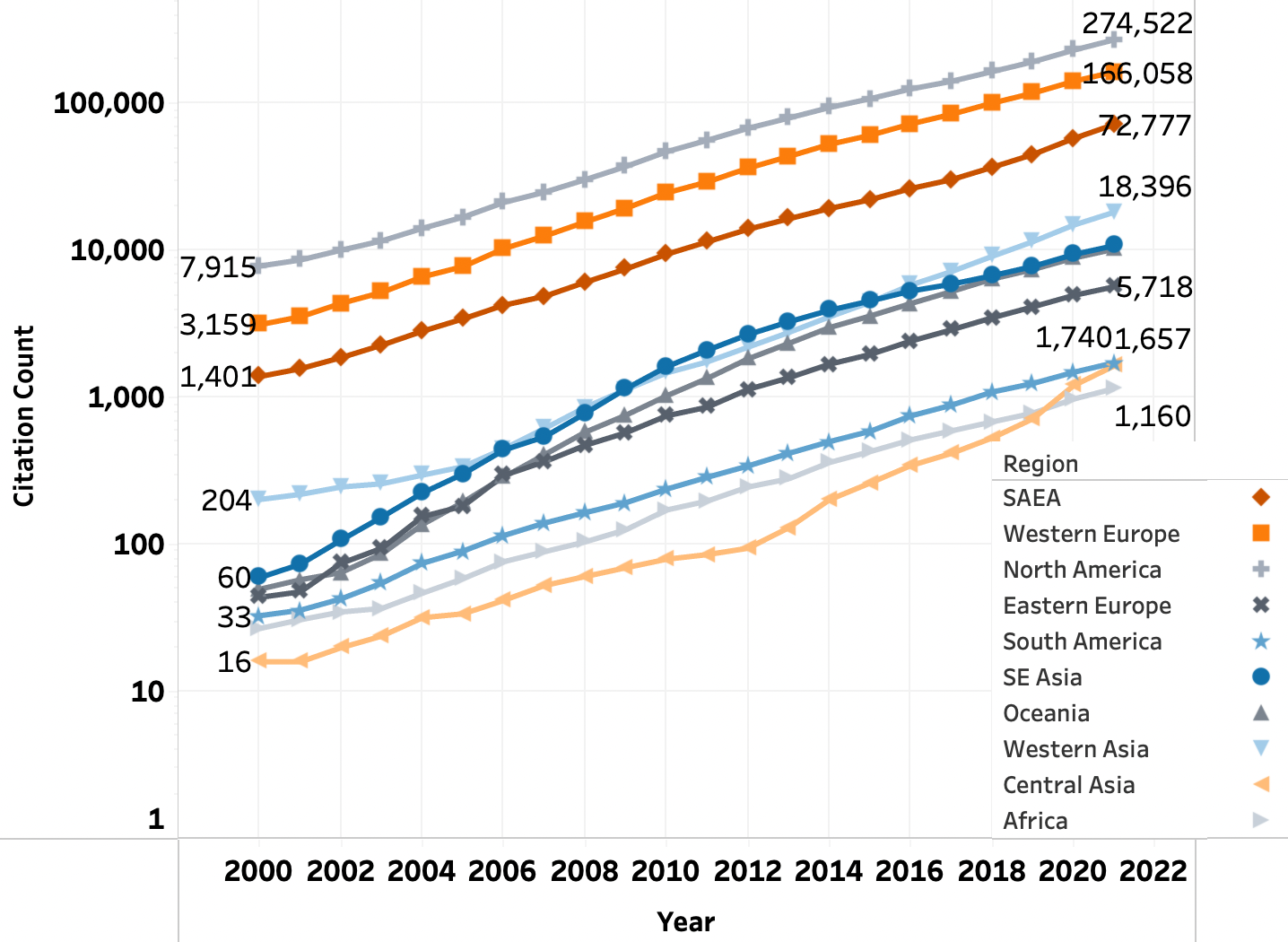}
    \caption{Citation count for each world region. \sm{Note the log scale for the y axis.} \texttt{SE Asia}: South Eastern Asia; \texttt{SAEA}: South Asia and Eastern Asia. }
    \label{fig:r0_citation_count}
\end{figure}

In Figure \ref{fig:r0_citation_count}, we plot the total citation count received by all the papers published by different regions of the world across across a span of $\sim$21 years from year 2000 to 2021.\footnote{We use shapes and colour-blind friendly palettes in all the figures to ensure that the figures are \textbf{\textit{colour-blind friendly}}. An \textit{\textbf{interactive version}} of all figures is accessible in Appendix \ref{sec:app_int_plot}.} 
Similarly, in Figure \ref{fig:time_citation}, we plot the mean-citation metric for top-10 countries across these years.
We restrict the analysis to just the top-10 countries for the sake of simplicity. 
\changejvj{ %
Also, note that in Figures \ref{fig:r0_citation_count} and \ref{fig:time_citation}, for each point in the x-axis, the corresponding point in the y-axis 
pertains to
the citation graph composed of papers published up till that year.}

\begin{figure}[t]
    \centering
    \includegraphics[width=1\columnwidth]{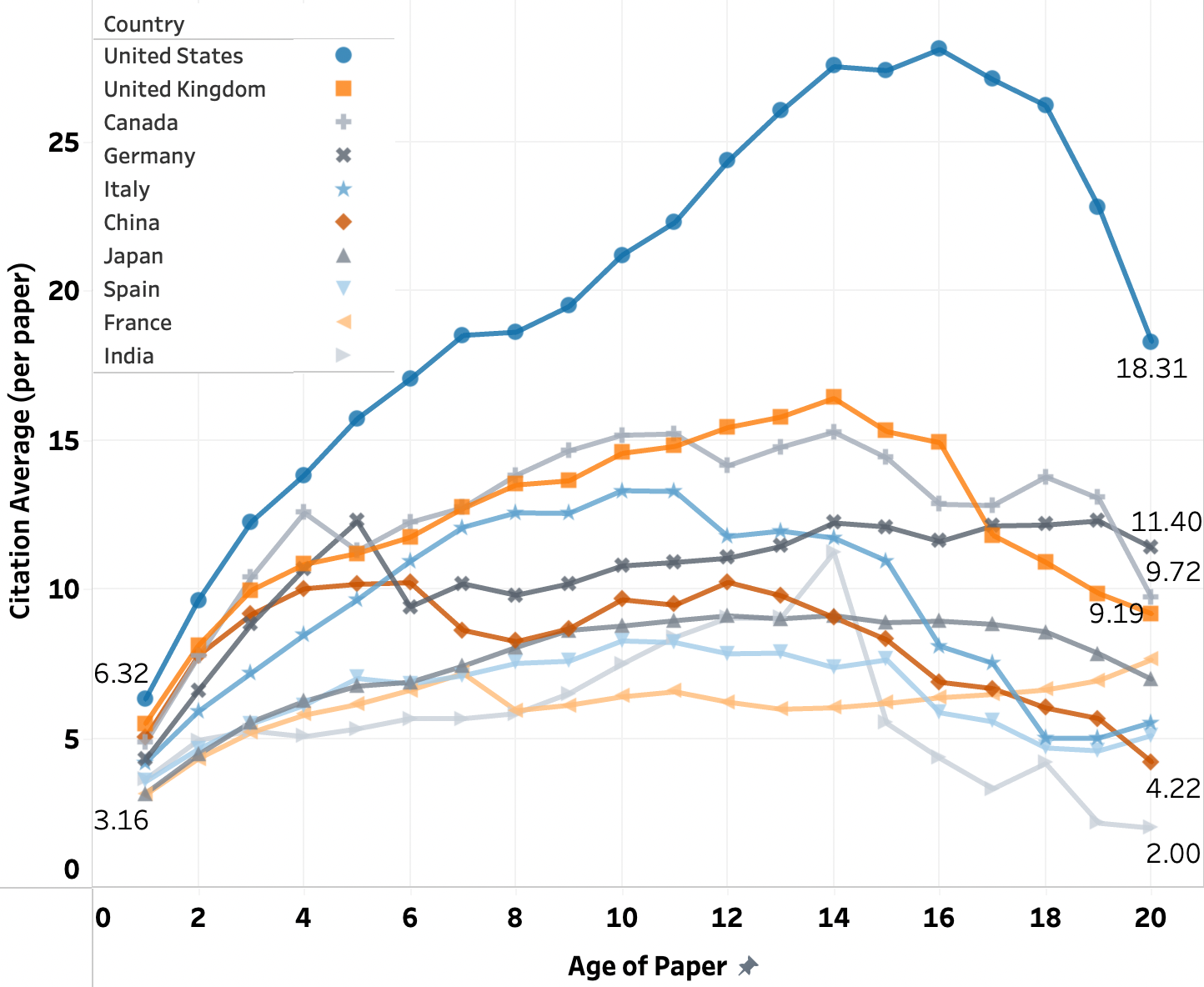}
    \caption{Variation of mean citations (per paper) across age of publication for the top-10 publishing countries.} 
    \label{fig:age_citation}
\end{figure}

\sm{To understand how the influence of a paper 
changes across the years after its publication,
for each country, we calculate the average number of citations its papers receive 
one year after publication, two years after publication, and so on.
We will refer to this time span as the age of a paper.
} 
We do so by building citation networks for each year from 2000 till 2021  and aggregate the number of papers and citations for each age-value (no. of years since the paper was published). Figure \ref{fig:age_citation} shows the plot of average citation vs.\@ age-of-paper.

\begin{figure}[t!]
    \centering
    \includegraphics[width=1\columnwidth]{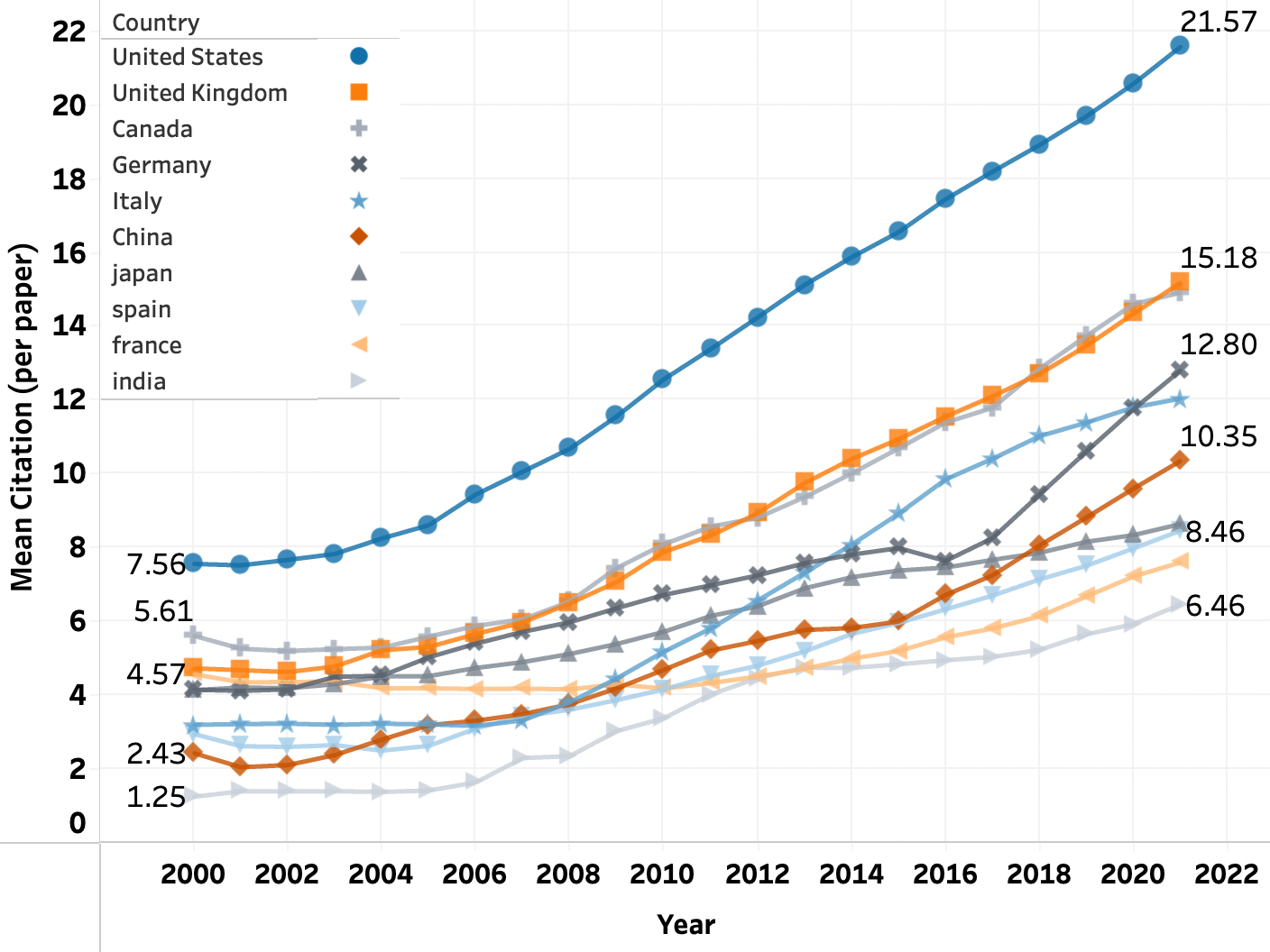}
    \caption{Variation of mean citations (per publication) across time for the top-10 publishing countries.}
    \label{fig:time_citation}
\end{figure}
\paragraph{Discussion} 
Figure \ref{fig:r0_citation_count} clearly depicts the huge disparities existing in citations received by different regions. 
\sm{Further, the trends have largely remained the same over the years, with the notable exception that central asia has made moderate gains since 2013, whereas the gains made by south east asia around the 2010s, have diminished since about 2012.}

From Figure \ref{fig:time_citation}, we make following observations: \textit{Firstly}, the plot shows that US (by a large margin), as well as, UK and Canada dominate the mean citation metric across the considered time-span. \textit{Secondly}, for these three countries as well as for Germany, the growth-rate of the mean-citation metric is remarkably higher than that of other countries. This growth-rate is indicated by the slope of the curve. \textit{Thirdly}, %
China and Germany \sm{have} shown a significant increase in growth rate over the past eight years. \textit{Overall}, this plot highlights that papers from certain countries have had markedly higher citations (on average) compared to other countries, 
across the years.
Additionally, we also note that the gap in mean-citation metric across the top-10 countries has \sm{more than doubled (increasing} from 6.31 to 15.11) in the past 
21 years. %
This also indicates that much of the   "influential" 
research in NLP from the past two decades is %
heavily concentrated in US, UK, and Canada.

We also explore median citation statistics 
in Figure \ref{fig:med_time_citation} (Appendix  \ref{sec:supplementary_plots}). In contrast to the mean, difference between medians is relatively smaller (for instance, 6 for US vs 3 for India in year 2021). 
This suggests that countries such as US, UK, and Canada have more very highly cited papers that push up their means. %
We also observe that the median has increased faster for countries like US, UK, and Canada compared to India, Spain, and France. Although the differences in medians are smaller, they are suggestive of marked citation disparity among the publication from different countries.

\sm{Figure \ref{fig:age_citation} 
shows that in the initial years after a paper's publication (1--2 years), the gap in mean-citation %
between the most and least cited} top-10 countries is $\sim$3.5, but it explodes  to $\sim$23 after 15 years. This suggest that, even among the top-publishing countries, the influence of some countries (like India and Spain) is surprisingly short-lived. Whereas, papers from US, UK, and Canada rapidly gain influence as the paper ages and their influence is long lasting.

\vspace{2mm}
\noindent
\textbf{Q3. How do countries cite each other? 
What inter-country citation patterns contribute to citation disparities?}\\

\begin{figure}[t]
    \centering
    \includegraphics[width=1\columnwidth]{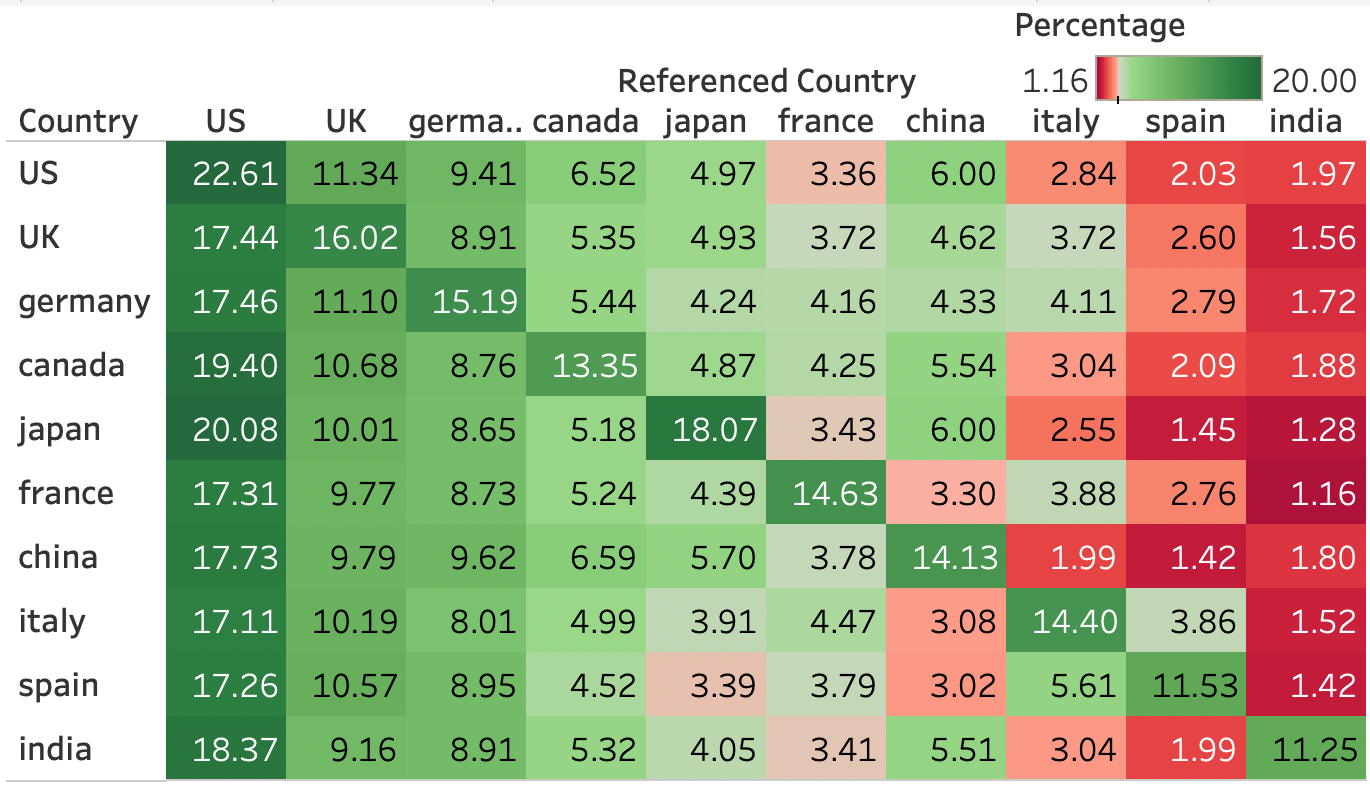}
    \caption{Heatmap depicting inter-country citation spread among top-10 publishing countries, until 2021.}
    \label{fig:r2_heatmap}
\end{figure}

\noindent
\textbf{A.} We analyze the following metric for all publications until year 2021: fraction of references from country-$j$ in an average paper from country-$k$. \sm{We will call it the {\it citation fraction}.}  Mathematically, %
it is defined as follows:

$$F(k, j)=\frac{\sum_{i \in P} \sum_{r \in R(i)} \mathbb{I}_{r \in j} \mathbb{I}_{i \in k}}{\sum_{i \in P} |R(i)| \mathbb{I}_{i \in k}}$$

\noindent where $R(i)$ indicates the set of all the references of paper-$i$, $\mathbb{I}_{r \in j}$ is $1$ if reference-$r$ (a paper) belongs to country-$j$ otherwise $0$, $P$ is the set of all the papers. Intuitively, $F(k, j)$ indicates the average fraction of references from country-$j$ in a paper from country-$k$. We compute the metric $F(k, j)$ from our citation graph with papers as nodes and references as edges. In Figure \ref{fig:r2_heatmap}, we plot $F(k,j)$ (times 100 or \%age) to study how a country contributes to the citation count of another country. Here, country-$k$ is shown in each row, with country-$j$ in each column. Moreover, we also study how country-$k$'s citation fractions' $F(k, j)$ dispersion has changed over years. To study this, we plot Gini-coefficient \citep{dorfman1979formula} for each each country for each year from 2000. Again, we restrict the plot to the top-10 publishing countries for simplicity.

\paragraph{Discussion} We make following observations from Figure \ref{fig:r2_heatmap}. \textit{Firstly}, for each country "self-citation" seems to be the highest contributor. In other words, a paper's most citations are from its' original country. %
\textit{Secondly}, all the countries cite US, UK, Germany, and Canada, which is shown by high density in columns corresponding to US (18.47\%), UK (10.86\% avg.), Germany (9.51\%) and Canada (6.25\%). These  convey that not only "self-citation" but significantly larger citation from all the other countries contribute to higher citation statistics for papers from some countries (like US, UK, Germany, and Canada).
Note that the sum of all the rows \sm{for a country} is $\sim$71. If all the referenced countries were plotted, then these row-sums would have resulted in a perfect 100. This means that $\sim$29\% of citations from the top-10 publishing countries are carried to the rest of the world. %
 
\begin{figure}[h!]
    \centering
    \includegraphics[width=1\columnwidth]{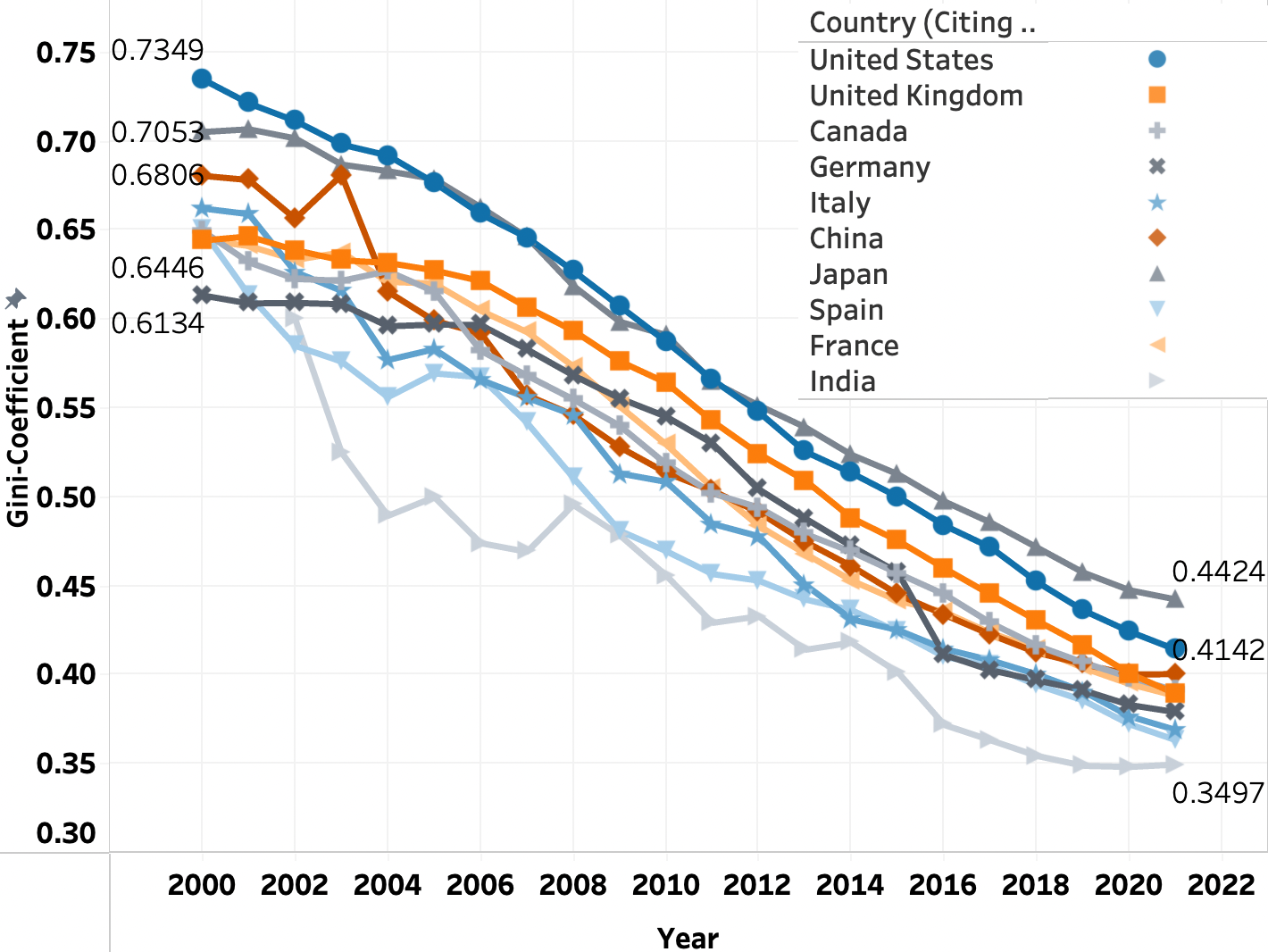}
    \caption{Gini-coefficient depicting dispersion of citation fractions among top-10 publishing countries.}
    \label{fig:r2_gini}
\end{figure}

In Figure \ref{fig:r2_gini}, we plot the Gini-coefficient for these citation fractions, to show how the spread of citations has changed over years. A decreasing value of Gini-coefficient indicates that the spread is becoming closer to equal (a proxy for research-inclusivity and equitable opportunities). The Gini-coefficient of citation-fraction drops almost linearly for all top-10 countries. Specifically, the average of Gini-coefficients 
drops from \sm{0.5985 to 0.3887}  from 2000 to 2021. This is encouraging as a decrease in Gini-coefficient indicates inclusivity in research.  
A linear interpolation, 
suggests that ideally it would still take $\sim$39 years to reach zero for these  Gini-coefficients, if the rate of change stays the same.

\vspace{2mm}
\noindent
\textbf{Q4. How does geographic location drive the collaboration between researchers? How has this changed over time?}\\

\begin{figure}[t]
    \centering
    \includegraphics[width=1\columnwidth,trim=0 0 0 0,clip]{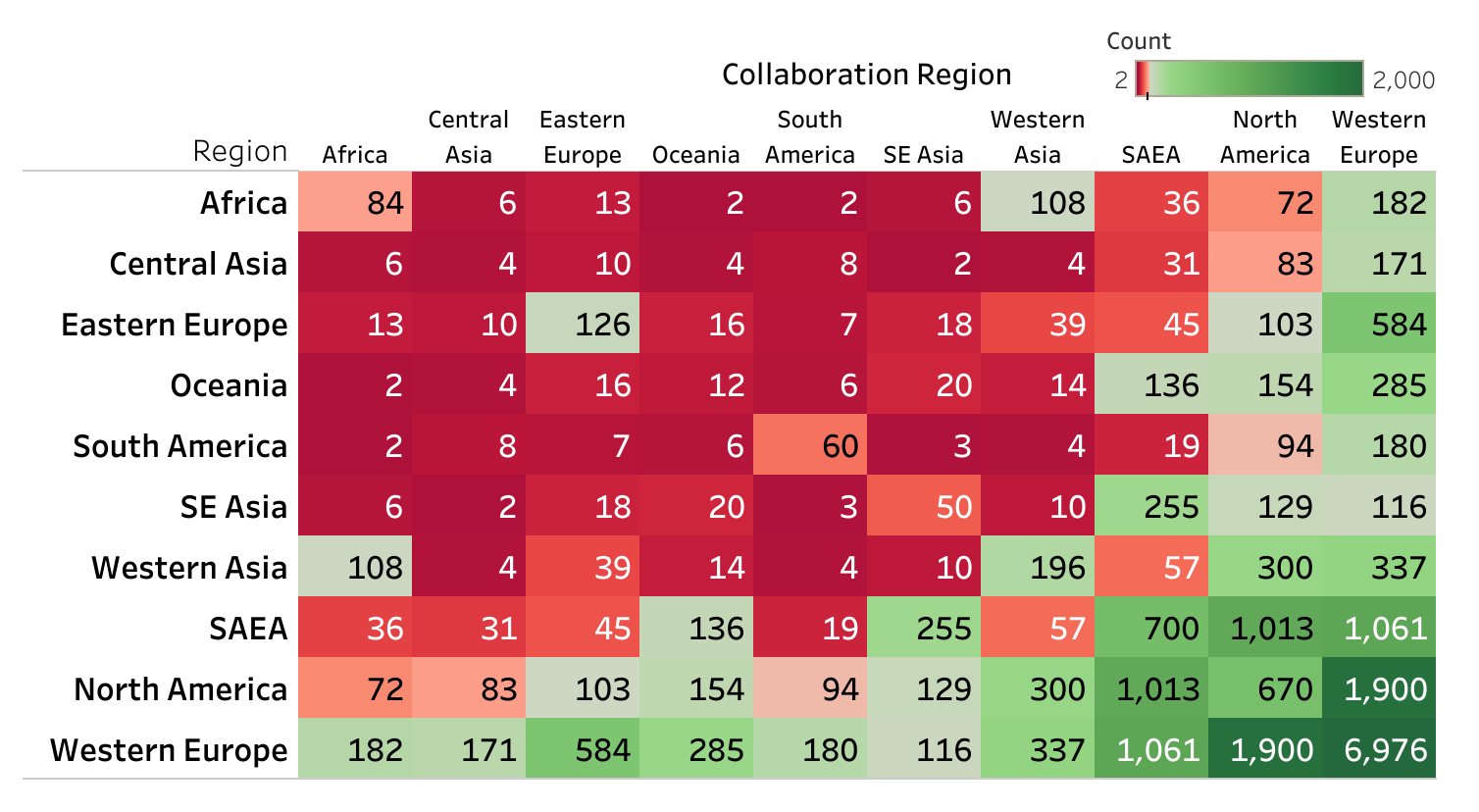}
    \caption{Heatmap indicating the collaboration between two regions with number of papers till 2021. \texttt{SE Asia}: South Eastern Asia; \texttt{SAEA}: South Asia and Eastern Asia.}
    \label{fig:r3_heatmap}
\end{figure}

\noindent
\textbf{A.} For this analysis, we choose to measure inter-country collaboration, i.e., only papers with authors from more than one country are considered. 
Figure \ref{fig:r3_heatmap} shows the heatmap with the count of the number of papers for all pairs of regions. This count indicates the number of papers belonging to the collaboration between the countries from a given region pair. Note that `\textit{collaboration between the same region}' includes only those papers having a collaboration between different countries in that region, and \sm{papers with all authors from the same country are excluded from this analysis.}

\begin{figure}[t]
    \centering
    \includegraphics[width=1\columnwidth,trim=0 0 70 30,clip]{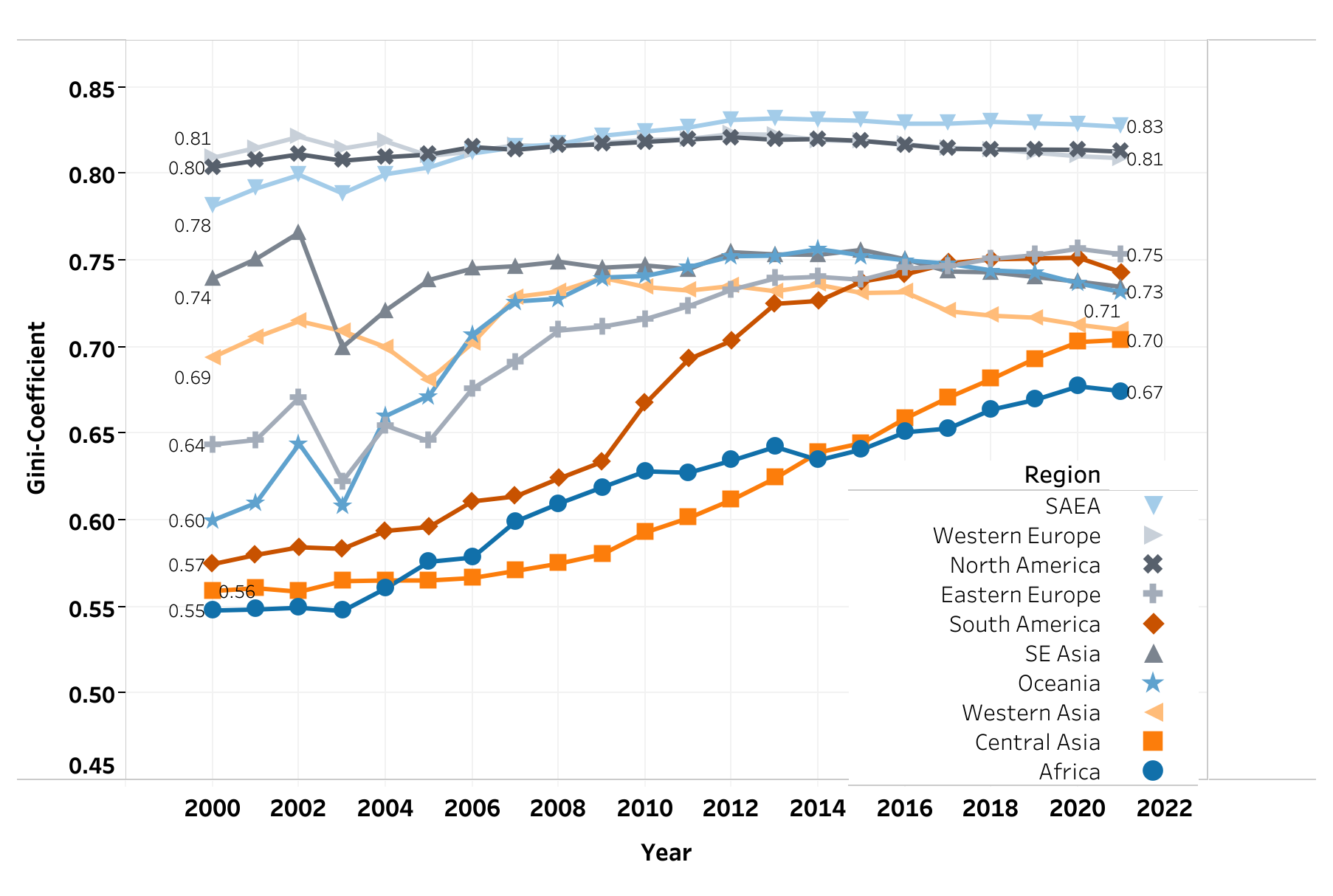} %
    \caption{Trend of Gini-coefficient depicting collaboration among different regions. \texttt{SE Asia}: South Eastern Asia \& \texttt{SAEA}: South Asia and Eastern Asia.}
    \label{fig:r3_gini}
\end{figure}

Again, we use the Gini coefficient to  quantitatively account for the collaboration between researchers across geographic locations.
This coefficient will be close to 0 if the regions collaborate roughly equally with other regions, and close to 1 for a very selective collaboration. We computed the Gini coefficient across time; and, to determine the coefficient for a given year, we only considered the papers published up till that year. Figure \ref{fig:r3_gini} shows the trend of Gini coefficient over years. %

\paragraph{Discussion} \sm{We can make three notable observations from the heatmap of  collaborations between continents (Figure \ref{fig:r3_heatmap}).}
Firstly, different regions of Europe collaborate well with each other, resulting in Europe having the highest inter-continent collaboration. Collaboration in North America can be mostly attributed to the collaboration between the United States and Canada. Secondly, North America has more collaboration with Europe (2,003) than Asia (1,525), indicating the preferential disparity of North American researchers towards Europe and Asia to some extent. Thirdly, South America and Africa have very few research collaborations with the rest of the world; within this limited collaboration, these regions have higher collaboration with Asia and Europe than with North America. \sm{These results highlight the dire need for a wider} research collaboration.
The Gini coefficient estimating the disparity in  collaboration across geographic regions is shown in Figure \ref{fig:r3_gini}. From 2000 to 2021, Gini coefficient for Africa rose from 0.37 to 0.53 and for South America, it rose from 0.41 to 0.61. 
Wide and meaningful collaboration is crucial for exchange of ideas and learning of new skills. Thus
this decreased collaboration 
with researchers from Africa and South America is concerning. 

\vspace{2mm}
\noindent
\textbf{Q5. How do the citation statistics of different countries vary across venues? %
}\\

\noindent
\textbf{A.} The venue of publication 
is believed to correlate somewhat to
the quality of the publication and its impact. 
Very roughly speaking, 
papers published in a venue would be of similar quality and thus may be expected to be cited similarly. 
For example, papers published in journals, \sm{such as} TACL, would be very similar to other papers in the venue, versus a paper published in a workshop. 
Here, we study citations of countries across venues to determine whether citation gaps exist even when one controls for the venue of publication.

\begin{figure}[t!]
    \centering
    \includegraphics[width=1\columnwidth,trim=0 0 60 20,clip]{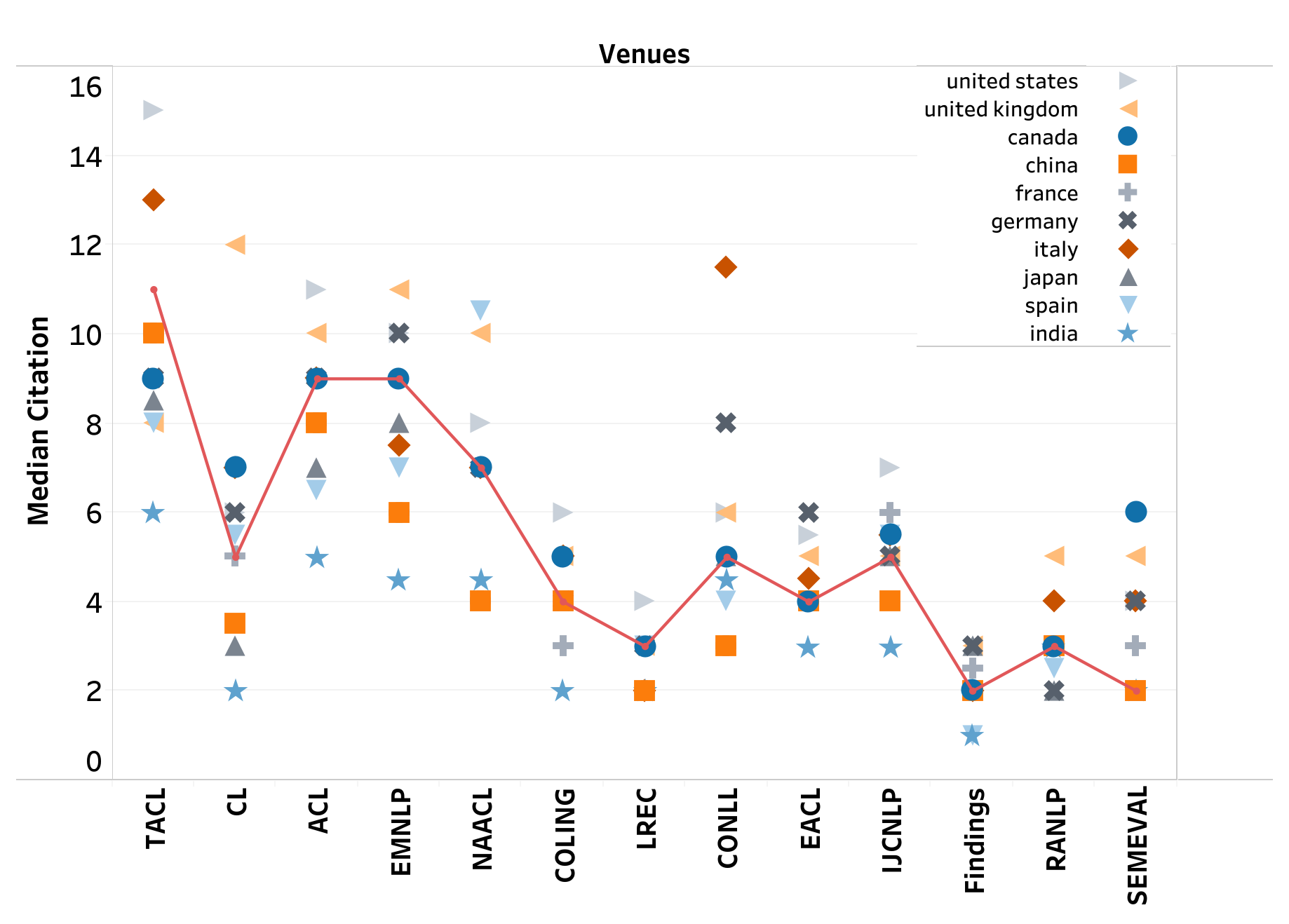}
    \caption{Median citation of papers published by different countries across different venues}
    \label{fig:r4_venue}
\end{figure}

We computed the median citation received by all papers published in a venue grouped by the country of publication. 
We choose the median as the measurement metric because it is less influenced by extremely large citation counts. Figure \ref{fig:r4_venue} shows the plot \sm{for the top-10 countries}.

\paragraph{Discussion} We observe that the United States is above this median line (indicated by a solid red line) for all 13 of 13 venues and the same can be said for the United Kingdom (10 of 13) and Germany(6 of 13). On the other hand, Japan is below the median line for 12 of 13 venues, and China and India are below the median line for all the venues. Even though the later countries publish fairly in all these venues, they receive significantly low citations compared to their peers. Also Figure \ref{fig:r4_venue} brings out the point that median citation for conferences like ACL, EMNLP, NAACL is significantly higher than Findings, RANLP, SEMEVAL. Overall this analysis shows the huge disparity in the citation received by the median paper from different countries published in the same venue. Thus geographic citation gaps exist even when we control for the venue of publication.

\vspace{2mm}
\noindent
\textbf{Q6. Is disparity in citation statistics consistent across research areas within NLP? Or is the gap simply because some countries work in areas that receive low numbers of citations (overall)?}\\

\begin{figure}[t]
    \centering
    \includegraphics[width=1\columnwidth,trim=0 0 60 20,clip]{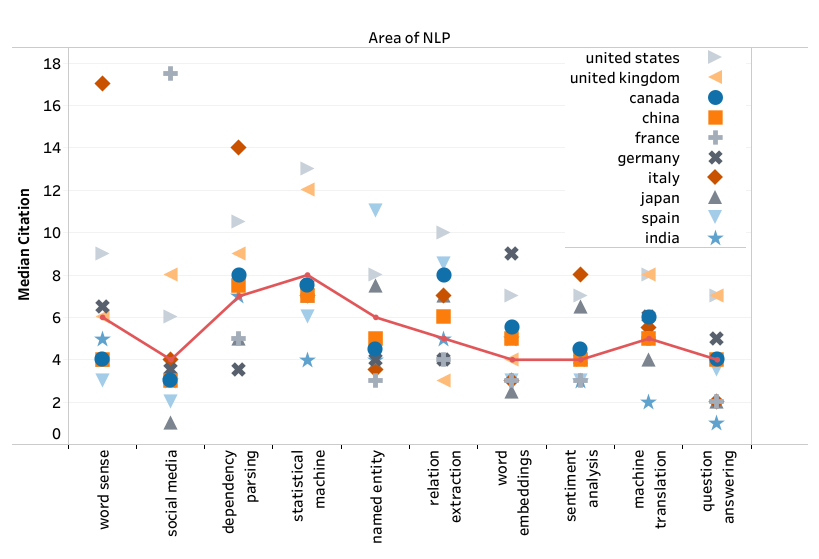}
    \caption{Median citation of papers published by different countries across different areas of NLP}
    \label{fig:r5_area}
\end{figure}

\noindent 
\textbf{A.} In order to sample papers from various sub-field/areas of NLP, we consider the word bigrams from the title of the paper. We follow the same approach used by \citet{mohammad-2020-gender}. %
We consider top-10 bigrams based on the number of papers sampled for each bigram, each representing an area of NLP. Figure  \ref{fig:r5_area} shows the median citation received by the publications from different countries across different areas of NLP. The median citation received by all papers published in an area is represented by a solid red line in Figure \ref{fig:r5_area}.

\noindent \textbf{Discussion:} %
Here we aim to understand whether low-cited countries work mostly in the areas that are less-cited. Conversely, we can also argue that highly cited countries mostly work in the areas that receive a high citation. From Figure \ref{fig:r5_area} we can see that the United States is above the median line for all 10 areas, the United Kingdom for 5 of 10. On the other hand, countries like China (3 of 10), India (1 of 10), and Japan (3 of 10) stay below the median line for the majority of areas. In order to fairly compare the difference in median citations among countries, we consider the number of papers published by any country in a particular area. Table \ref{tab:r5_comparison} compares the publication count of the United States and China for 3 areas of NLP and their median citation. China has a fairly similar number of publications for these areas, but the median citation is significantly lower than the United States. This highlights the disparity in citations received by papers based on their geographic location. 

\begin{table}[t!]
\begin{center}
\setlength{\tabcolsep}{5pt}%
\scalebox{0.75}{
\begin{tabular}{lrrrr}
\toprule
\multirow{2}{*}{\textbf{Area}}
& \multicolumn{ 2}{c}{\textbf{United States}}
& \multicolumn{ 2}{c}{\textbf{China}} \\
\cmidrule(lr){2-3} \cmidrule(lr){4-5}
& \textbf{Count}
& \textbf{Median}
& \textbf{Count} 
& \textbf{Median} \\

\midrule

Sentiment Analysis & 82 &  7.0 & 65 & 4.0 \\
Dependency parsing & 96 & 10.5 & 68 & 7.5 \\
Relation Extraction & 100 & 10.0 & 81 & 6.0 \\

\bottomrule 
\end{tabular}
}
\end{center}
\caption{\label{tab:r5_comparison} Total papers and mean citation statistics for US and China across areas of research.
}
\end{table}

\section{Related Work}

Bias in citation networks is a long standing problem that arises by a combination of a myriad number of factors. \citet{tahamtan2016factors} categorised these factors into three buckets: firstly, paper related factors which involves paper quality \citep{buela2010analysis}, field of study \citep{costas2009scaling}, novelty, length of paper \citep{antoniou2015bibliometric, falagas2013impact}; secondly, journal/conference related factors, like impact factor \citep{callaham2002journal}, language of publication \citep{lira2013influence};  and thirdly, author related factors, like number of authors  \citep{della2008multi, bosquet2013academics}, collaboration \citep{nomaler2013more}, self-citation \citep{costas2010self} author’s reputation \citep{collet2014does}, affiliation \citep{sin2011international}, gender, race and age \citep{ayres2000determinants, leimu2005determines}.

\citet{nielsen2021global} analyzed a few subfields of science to show that the concentration of citation by the top 1\% of the authors increased from 14 to 21\% between 2000 and 2015 and that the Gini coefficient for citation imbalance has risen from 0.65 to 0.70. They also observed that these top 1\% authors mostly reside in Western Europe and Australasia. Similarly, \citet{lee2010author} and \citet{pasterkamp2007citation} presented a similar hypothesis that researchers from USA and UK are cited much more than rest of the world. This finding is well established across different disciplines of science. Using correlation analysis, \citet{lou2015does} have shown that there is a high negative correlation between the author's affiliation and uncitedness. They also show that along with the affiliation, the venue of publication greatly impacts the citation received by any paper \citep{callaham2002journal}.
\citet{schluter2018glass} shows, using a mentor-mentee network, that there is a significant time gap for the female researchers to acquire the mentorship status. \citet{vogel-jurafsky-2012-said, mohammad-2020-gender} measures the gender bias existing in the NLP community by considering papers published in ACL Anthology. However, \citet{mohammad-2020-gender} considers a larger domain to examine the gender bias specifically from the point of view of female first and last authors. 

In the broader science community, extensive research has shown the existence of gender bias in citation network and its impact on the career trajectory of female researchers \citep{chatterjee2021gender, llorens2021gender}. Although research is done in quantifying the citation disparity across geographic location for other fields \citep{paris1998region, nishioka2022does}, not many studies have systematically studied geographic gap in NLP, and our work makes an original contribution in analyzing this disparity at scale in the NLP community.

\section{Conclusion}
We compiled a large dataset of meta-information associated with NLP papers,
that includes the country name associated with an author's affiliated institution, number of citations, and the papers cited by each paper. 
We use this dataset to systematically examine citation disparities across geographic regions. 
We show that substantial geographic disparities exist, even when controlling for factors such as venue of publication.
Authors from North America and Western Europe have published 65\% of the total number of papers; whereas, authors from Africa, South America, and South-Eastern Asia (combined) have published only 3.7\% of the total papers. 
We also find markedly lower levels of cross-region author--author collaboration across the two sets of regions. 
Even among the top-ten publishing countries, we find marked differences. Very little research from countries such as India, Spain, Italy, and China is cited by their peers from the top-five publishing countries. 

\changemrungta{Citation gaps across genders and geographic regions occur due to various complex, intersectional, and structural reasons. This paper does not explore the reasons behind them. However, \citet{pan2012world, skopec2020role, wuestman2019geography} have examined some of the causes 
for geographic citation gaps in scientific literature
such as self citation, co-location of institutions, and national funding for research \& development.
The above reasons are not exhaustive and determining the degree of impact of each reason is extremely difficult. Further,
the reasons can be different across different regions of the world. We hope that future qualitative research sheds greater light on these complexities. Understanding them is crucial for greater inclusion in research.}
We also hope that such work will engender concrete actions to address citation gaps. 
Initiatives taken by the NLP community toward diversity, inclusion, and collaboration, especially actions that encourage local NLP research across the globe, can make meaningful change and help address geographic inequities. 

\section{Ethics Statement}
Since the attribution of paper to countries is done through an automatic process, \sm{it is possible that
a small number of attributions} might suffer from noisy labels. 
Further, the only inferences drawn in this work are at aggregate-level, and not about individual papers or authors.

\section{Limitations}
\label{"sec:limit"}

ACL Anthology does not provide the affiliation (or some location information) associated with the authors. Since our work majorly relies on authors' location, we came up with a heuristic approach (described in Section \ref{"subsec:country_extract"}) to automatically identify the country of the authors' associated with each research paper. Additionally, we also analyzed the correctness of automatically identified country tags. In our analysis, we found that for 84.5\% of the papers country tags were perfectly retrieved. For the remaining unmatched papers, the algorithm either missed one country from the list or was unable to find any country tag for the paper. Since we use an exhaustive list of countries and universities, these retrieval errors are not country-specific. But rather these errors are specific to the formatting of the research papers, which is dependent on the publication venue. 
\sm{In order to facilitate future research, we suggest that ACL should make location-associated meta-information of the authors publicly available. We also suggest that conferences and journals should publicly report summary statistics of geographic diversity of accepted papers.}

\sm{In this work, we only used the papers published in the ACL Anthology.
These are usually papers from large international conferences, and always written in English.
However, it should be noted that there exist vibrant local communities 
that publish at local venues, often in non-English languages. Furthermore, NLP research is also published at other non-ACL global venues, such as AAAI, ICLR, ICML, WWW. The conclusions drawn from our experiments therefore apply to international English NLP conferences and journals, and further work is needed to explore the landscape of local sub-communities across the world.}

\changemrungta{Citations are heterogeneous and can be categorized in different ways. For example, semantic scholar categorizes citations into background, method, and result citations. In the current work, we do not distinguish between types of citations.}
The first-author position in a paper 
is usually reserved for the researcher that has done the most work and writing. 
The last author position is often reserved for the most senior or mentoring researcher.
However, this work does not explore the distributions of these positions by region.

\sm{Finally, even though this work only looks at published papers, future work can also examine rejection rates by geographic location (if that information becomes accessible). See \citet{church_2020} for a discussion on improving reviewing, and how organizers of early EMNLP conferences (in the early 2000s) actively accepted papers from Asia at time when that was less common. \citet{church_2020} argue that this was one of the reasons for the success of EMNLP and its reputation as a conference that is more accepting of new and diverse ideas.}

\bibliography{anthology,custom}
\bibliographystyle{acl_natbib}

\appendix

\section*{Appendix}
\label{sec:appendix}

\section{Supplementary Plot}
\label{sec:supplementary_plots}

\begin{figure}[h!]
    \centering
    \includegraphics[width=1\columnwidth]{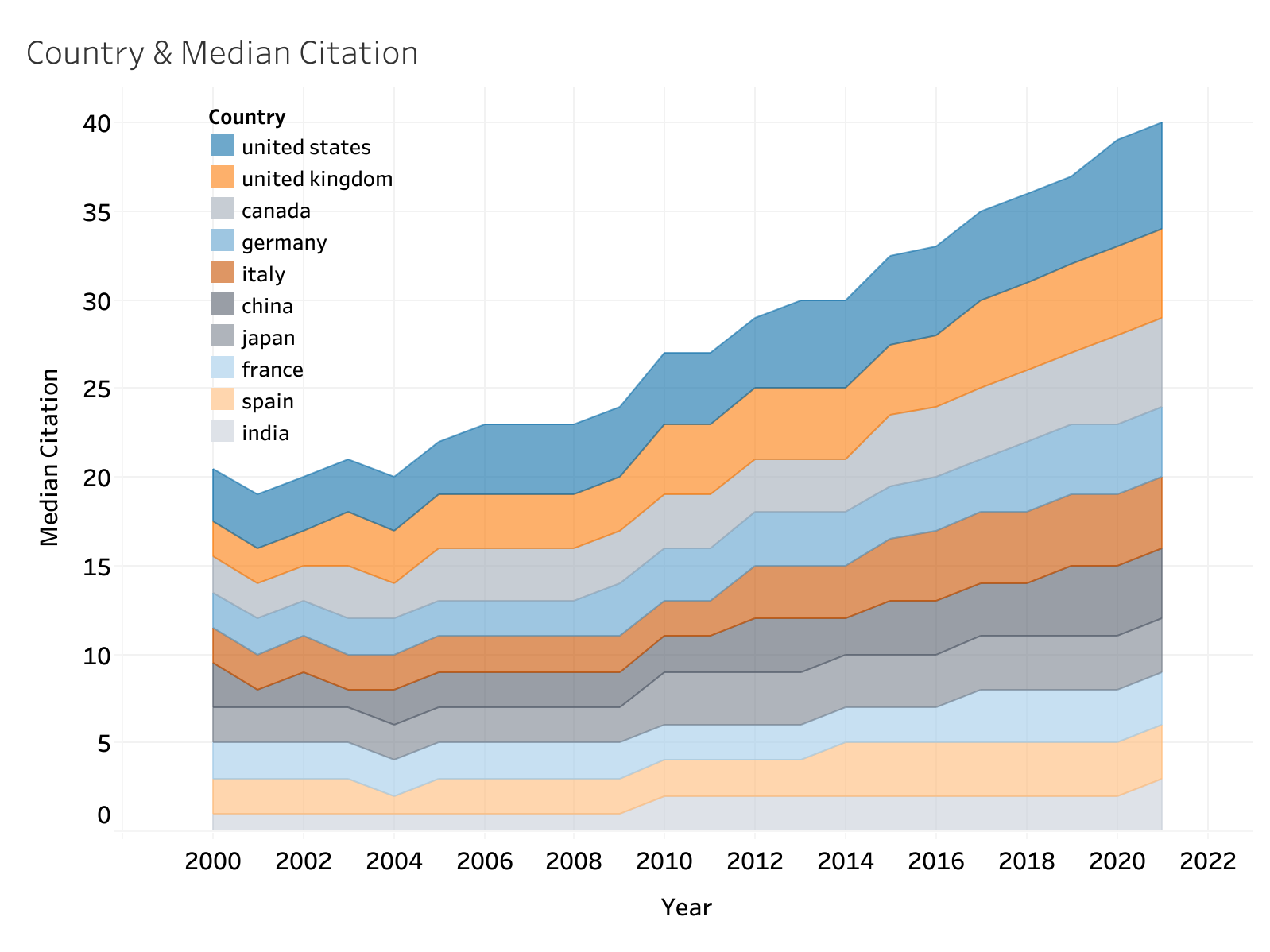}
    \caption{Variation of median citations (of all publication) across time for the top-10 publishing countries. (height of strip denotes median value)}
    \label{fig:med_time_citation} 
\end{figure}

Variation of median citations (of all publication) across time for the top-10 publishing countries is shown in Figure \ref{fig:med_time_citation}.

\section{Interactive Plots}
\label{sec:app_int_plot}
An interactive version for all the plots presented in this work can be accessed from following Tableau Public profile: \url{https://public.tableau.com/app/profile/gdnlp}

\section{Country Information Extraction}

\label{"app:loc_extraction"}
Below we describe the methodology used to extract country tag for each paper.
\begin{enumerate}
    \item Text extraction: We used scipdf-parser to extract the header information from the pdf document of the given research paper.
    \item Extracting country and organization tag: Text from the header is retrieved in JSON format where a map between country and organization is established by this tool. For some papers, the tool is not able to match the country tag to the appropriate country either because it is unable to parse it correctly or because the information is not present at all. Therefore, we use the value present in the organization tag to map the paper to the respective country. 
    \item Preprocessing: We remove all the non-alphanumeric characters from the extracted text from the last step. This is done to clean the text and perform the next step of country extraction easier.
    \item University list: Papers for which the country is not available, we perform matching the organization name to its affiliated country. We created a list of all universities of the list by following some earlier works. We mined this page\footnote{\url{https://www.4icu.org/reviews/index2.htm}} that contains a consolidated list of universities all across the globe. We augment this list with the University list from this link\footnote{\url{https://raw.githubusercontent.com/endSly/world-universities-csv/master/world-universities.csv}} as well. Thus, we finally got a list of around 18k universities and their corresponding country. Next, text matching is performed on the extracted organization name with the university list to get the country associated with the paper.  
    \item Country List: %
    In a few cases the tool was unable to extract the country information even though it was present. Therefore, to solve this we curate a list of all countries across the globe. We perform direct text matching of the extracted text with the country list and map the paper to its corresponding country.
\end{enumerate}

\section{Country to Region Mapping}
\label{app:map_count_to_region}
We aggregate information from different countries into regions. This country to region mapping is chosen so as the countries included in a region are in geographical vicinity. \changemrungta{In this mapping, we have considered different variations of country names in the affiliations. We consider these multiple variations so as to retrieve as many country mentions as possible from each paper.}
\begin{itemize}
    \item \textbf{South Asia and Eastern Asia (SAEA)}: afghanistan; bangladesh; bhutan; china; hong kong; india; iran; japan; korea; maldives; mongolia; nepal; north korea; pakistan; sri lanka; taiwan; vietnam 
    
    \item \textbf{Oceania}: australia; fiji; kiribati; marshall islands; micronesia; nauru; new zealand; palau; papua new guinea; samoa; solomon islands; tonga; tuvalu; vanuatu 
    
    \item \textbf{South Eastern Asia (SE Asia)}: brunei; cambodia; east timor; indonesia; malaysia; myanmar; philippines; singapore; thailand 
    
    \item \textbf{Eastern Europe}: belarus; bulgaria; cz; czech republic; czechia; hungary; moldova; poland; romania; russian federation; slovakia; ukraine 
    
    \item \textbf{Africa}: algeria; angola; benin; botswana; burkina; burundi; cameroon; cape verde; central african republic; chad; comoros; congo; congo, democratic republic of; djibouti; egypt; equatorial guinea; eritrea; ethiopia; gabon; gambia; ghana; guinea; guinea-bissau; ivory coast; kenya; lesotho; liberia; libya; madagascar; malawi; mali; mauritania; mauritius; morocco; mozambique; namibia; niger; nigeria; rwanda; sao tome and principe; senegal; seychelles; sierra leone; somalia; south africa; south sudan; sudan; swaziland; tanzania; togo; tunisia; uganda; zambia; zimbabwe 
    
    \item \textbf{North America}: antigua and barbuda; bahamas; barbados; belize; canada; costa rica; cuba; dominica; dominican republic; el salvador; grenada; guatemala; haiti; honduras; jamaica; mexico; nicaragua; panama; saint kitts and nevis; saint lucia; saint martin; saint vincent and the grenadines; trinidad and tobago; united states; us 
    
    \item \textbf{South America}: argentina; bolivia; brazil; chile; colombia; ecuador; guyana; paraguay; peru; suriname; uruguay; venezuela 
    
    \item \textbf{Central Asia}: kazakhstan; kyrgyzstan; russia; tajikistan; turkmenistan; uzbekistan 
    
    \item \textbf{Western Asia}: armenia; azerbaijan; bahrain; cyprus; georgia; iraq; israel; jordan; kuwait; lebanon; oman; qatar; saudi arabia; syria; turkey; united arab emirates; yemen 
    
    \item \textbf{Western Europe}: albania; andorra; austria; belgium; bosnia and herzegovina; croatia; denmark; estonia; finland; france; germany; greece; iceland; ireland; italy; latvia; liechtenstein; lithuania; luxembourg; macedonia; malta; monaco; montenegro; netherlands; norway; portugal; san marino; serbia; slovenia; spain; sweden; switzerland; united kingdom; vatican city 
\end{itemize}

\end{document}